\documentclass[lettersize,journal]{IEEEtran}
\usepackage{amsmath,amsfonts, amssymb}
\usepackage{algorithmic}
\usepackage{algorithm}
\usepackage{array}
\usepackage[caption=false,font=normalsize,labelfont=sf,textfont=sf]{subfig}
\usepackage{textcomp}
\usepackage{stfloats}
\usepackage{url}
\usepackage{verbatim}
\usepackage{graphicx}
\usepackage{cite}
\usepackage[dvipsnames]{xcolor}
\begin{document}

\title{Block-as-Domain Adaptation for Workload Prediction from fNIRS Data}

\author{\IEEEauthorblockN{Jiyang Wang,
Ayse Altay, Senem Velipasalar\\}
\IEEEauthorblockA{EECS,
Syracuse University\\
jwang127@syr.edu,
aaltay@syr.edu,
svelipas@syr.edu}}

\markboth{Journal of \LaTeX\ Class Files,~Vol.~14, No.~8, August~2021}%
{Shell \MakeLowercase{\textit{et al.}}: A Sample Article Using IEEEtran.cls for IEEE Journals}

\IEEEpubid{0000--0000/00\$00.00~\copyright~2021 IEEE}

\maketitle

\begin{abstract}
Functional near-infrared spectroscopy (fNIRS) is a non-intrusive way to measure cortical hemodynamic activity. Predicting cognitive workload from fNIRS data has taken on a diffuse set of methods. To be applicable in real-world settings, models are needed, which can perform well across different sessions as well as different subjects. However, most existing works assume that training and testing data come from the same subjects and/or cannot generalize well across never-before-seen subjects. 
Additional challenges imposed by fNIRS data include the high variations in inter-subject fNIRS data and also in intra-subject data collected across different blocks of sessions.
To address these issues, 
we propose an effective method, referred to as the class-aware-block-aware domain adaptation (CABA-DA) which explicitly minimize intra-session variance by viewing different blocks from the same subject same session as different domains. We minimize the intra-class domain discrepancy and maximize the inter-class domain discrepancy accordingly. 
In addition, we propose an MLPMixer-based model for cognitive load classification. Experimental results demonstrate the proposed model has better performance compared with three different baseline models on three public-available datasets of cognitive workload. Two of them are collected from n-back tasks and one of them is from finger tapping. 
From our experiments, we also show the proposed contrastive learning method can also improve baseline models we compared with.
\end{abstract}


\section{Introduction}
The level of cognitive workload (CWL) that a human experiences can affect their performance in human computer interaction (HCI) tasks. A high CWL can lead to human error, such as task shedding and frustration, while a low CWL can cause boredom and complacency \cite{mandrick2016neural}. A real-world autonomous system that can dynamically adjust and assign tasks based on the CWLs of different subjects can enhance both the efficiency and the satisfaction of the human participants. Such a system needs a model that can generalize well to new subjects that have not been seen before.
Neuroscience researchers often use functional magnetic resonance imaging (fMRI) to study the brain activity of humans, because it has high spatial and temporal resolution~\cite{dimoka2012conduct}. However, fMRI is not very suitable for human computer interaction (HCI) research, because it is costly and prone to motion artifacts. Participants have to remain still while the data is collected. Therefore, HCI researchers have explored other methods to measure brain activity, such as electroencephalography (EEG) and functional Near-Infrared Spectroscopy (fNIRS). EEG measures the electrical potential on the scalp that is generated by neural activation on the surface of the brain. It uses electrodes that are attached to the scalp. fNIRS uses near-infrared light that can pass through the scalp and skull and reach the cortical surface of the head. It measures the light that is reflected back from the brain tissue. The change in light intensity indicates the changes in oxy- and deoxy-hemoglobin concentration~\cite{chance1993cognition}. Some tasks and experiments may involve movement of participants, such as typing on a keyboard. fNIRS is more resistant to electrical noise and motion-based muscle activity artifacts than EEG~\cite{shin2018simultaneous}.

\IEEEpubidadjcol
In this work, we study the problem of classification of cognitive workload levels from fNIRS data.~Methods have been proposed to employ deep learning for CWL classification. However, many of those approaches~\cite{saadati2021application, saadati2019mental, saadati2019convolutional, saadati2019multimodal, mughal2021fnirs} train models per individual, and perform training and testing on the same subject. While these approaches have explored the potential of using deep learning on brain data, many challenges and issues have remained that need to be addressed. These issues include overfitting and inflated accuracy rates. When these models are tested on a new subject or even on the same subject during a different measurement session, their performance degrades significantly. This degradation may be in part due to the assumptions made by deep learning operations, and the high reliance on Convolutional Neural Networks (CNN)-based models~\cite{saadati2021application, asgher2020classification, mughal2021fnirs}.

The parameter sharing aspect~\cite{alzubaidi2021review} of CNNs is based on the assumption that a valid patch of weights working for one position also works for other regions. For instance, in image classification, if there is a filter that can extract features of an object, then,  no matter where the object is in the image, that filter will still be useful. Thanks to parameter sharing, CNNs have significantly less trainable parameters compared to multi-layer perceptrons (MLP). Although CNNs are considered to be invariant to small level of transformations, later studies~\cite{azulay2018deep, zhang2019making} have shown that this invariance assumption does not always hold. Moreover, there are some situations for which parameter sharing is not suitable. 

Prior fMRI literature on working memory load and selective attention has shown that the middle temporal gyrus is a region involved in the coordination between working memory and directed attention~\cite{sabri2014neural}. In addition, human brain is highly interconnected, and some brain regions can work together in cognitive processing. Thus, the assumption of CNNs that a patch of weights learnt from one spatial or temporal position can be used in another region does not always hold.

On the other hand, the number of probes in fNIRS devices and the corresponding fNIRS channels is usually less than 52~\cite{wang2021taking, huang2021tufts, shin2018simultaneous}. Hence, the processing of fNIRS data is not as computationally expensive as processing of images or videos. Motivated by these, instead of using CNNs, we propose an MLP-based method. More specifically, we employ an MLPMixer~\cite{tolstikhin2021mlp} architecture and adapt it to fNIRS data.~MLPMixer is a general backbone originally proposed for the image classification task, wherein the input is a sequence of non-overlapping image patches. Each patch is projected to a desired hidden dimension as a token. Two types of MLP blocks have been used in MLPMixer; one maps tokens along the sequence of tokens dimension, another one maps tokens along the channel (feature depth) dimension. In contrast, with fNIRS data, instead of divide input into patches, we project the entire spatial dimension to a token, due to the limited resolution of brain signal data. Thus in our version, the mix is happening on temporal axis.        
\begin{figure}[h]
    \centering
    \includegraphics[width=\linewidth]{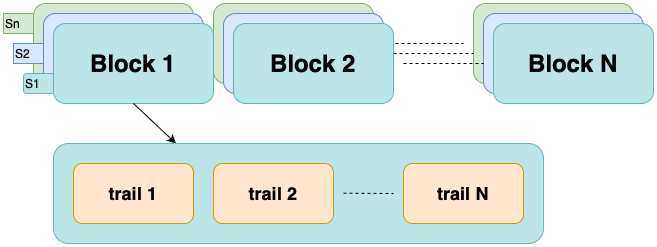}
    \caption{Commonly used experiment design when collecting fNIRS data. Different colors (best view in color) show different sessions at the top part, and the orange color indicating trails.}
    \label{fig:data_blocks}
\end{figure}

fNIRS data also introduces additional challenges for autonomous processing tasks, including CWL classification. More specifically, the high variations in inter-subject fNIRS data (when the same task is being performed) as well as in intra-subject data captured during different sessions need to be addressed. 
Block-wise experiments are a common method for collecting fNIRS data~\cite{shin2018simultaneous, huang2021tufts}. In this type of experiment, the participant performs a series  trails, that are repeated in blocks and in each trail, the participants are required to do one specific task. Fig.~\ref{fig:data_blocks} illustrates the structure of block-wise experiments for one participant.
The inter-subject variance arises due to the differences in type and style of hair, skin, skull and brain structure. Intra-subject variances can occur due to differences in sensor placement across different sessions. Even within the same session, head motion, body motion or noise due to the light source or electronics can cause difficulties~\cite{huang2021tufts, yucel2021best}. 
To address this problem, researchers~\cite{lyu2021domain, huang2021tufts, yucel2021best} have viewed data from different sessions of the same subject or from different subjects as data from different domains, and used domain adaptation (DA) to align the data. However, in their work, some channels with low signal-to-noise (SNR) need to be removed before the alignment process. 
Yet, dropping of channels can cause loss of useful information, and thus limit the performance, especially considering that the amount of data and the number of channels is already limited to begin with. Zhong et.al~\cite{zhong2020eeg} view different subject as different domains to solve the difference of distribution between different subjects. In an earlier work~\cite{wang2021taking}, we presented a self-supervised method, which augments the data by arranging the controlled rest and task windows in different orders. This approach trains the deep learning model to classify the combinations of different orders and cognitive levels at the same time, and provides performance improvement on different types of cognitive tasks. However, the reliance on controlled rest data for training limits its application when no form of controlled rest signal is available in the data.

We conducted preliminary experiments to train the DeepConv~\cite{schirrmeister2017deep} model on the TUBerlin data under different scenarios. We split the data by trial, block, session, and subject, and evaluated the model’s ability to classify the data from the three-way n-back (0- vs. 1- vs. 2-back) task. We describe this experiment in detail in Ablation Study at Sec.\ref{subsec:impact_of_intra_variance} and present the results in Tab.~\ref{tab:ab1}. The results show that the performance dropped by a similar amount when we changed the split from trial to block and from block to session. This implies that the block and session factors introduced similar levels of intra-class variance to the data. Previous studies focus on intra-class variance introduced by the difference of subjects and sessions and didn't study this intra-class variance in blocks~\cite{huang2021tufts, yucel2021best, lyu2021domain}. 

In order to address the aforementioned issues of high variance of inter-subject and intra-subject data, and eliminate the reliance on control rest data, we propose to view different different blocks from same/different subjects as different domains and use contrastive learning to align features extracted from different subjects for the same class.  
As mentioned above, previous works~\cite{lyu2021domain, zhong2020eeg} treat data from different subjects as different domains and apply domain adaptation to align samples and help deep learning models generalize. However, fNIRS data is usually collected through multiple separated sessions, and variation can exist even for the same subject across different sessions. Thus, different from existing works, we propose a \textit{block-aware loss} to also align samples across different blocks to improve the generalization of models. The beauty of this alignment is it also align the features from different sessions.

The main contributions of this paper 
include the following: 
\begin{itemize}
    \item We propose block discrepancy term to measure the differences between blocks within the same session. 
    \item We propose a contrastive learning method that aims to align the samples from different blocks that belong to the same class.
    \item We propose a new classification model based on the MLPMixer by adding mixing along the temporal dimension 
    \item We show that our proposed classification model outperforms three different baselines
    \item we show that contrastive learning modules can improve the performance of other deep learning models for fNIRS data.
\end{itemize}

The rest of this paper is organized as follows: Sec.~\ref{sec:related_work} provides an overview of the related work on non-invasive brain data followed by a review of the recent deep learning methods used in HCI. The details of our proposed method are presented in Sec.~\ref{sec:method}. The description of the datasets and the details of training and testing splits are provided in Sec.~\ref{sec:experiments} together with the experimental results and the related discussion. The paper is concluded in Sec.~\ref{sec:summ} with a summary and future work directions. 

\section{Related Work}
\label{sec:related_work}
Predicting workload (WL) has been of interest since as early as 1908~\cite{yerkes1908relation}. More recently, new approaches have been proposed to leverage machine learning for cognitive workload (CWL) classification.  
Several works~\cite{saadati2019convolutional, mughal2021fnirs} focus on per-participant model training due to high inter-subject as well as intra-subject cross-session variances of fNIRS data. Although providing good results, per-participant model training is not practical and not generalizable for real-world applications. As discussed above, more generalizable methods are needed, which can perform well on new data captured from never-before-seen participants. In this section, we first provide an overview of the work on non-invasive brain data, and summarize recent deep learning-based approaches that have been proposed to analyze brain signals.

\subsection{Non-invasive brain data}
\label{subsec:noninvasive}
As mentioned above, although functional magnetic resonance imaging (fMRI) is very commonly used thanks to its high spatial and temporal resolution~\cite{dimoka2012conduct}, it has limitations as a research tool. In addition to being expensive, fMRI is very sensitive to motion artifacts.%
Participants need to keep still during data collection, since their movements can cause issues. For these reasons, researchers in the HCI domain have also focused on electroencephalography (EEG)~\cite{lawhern2018eegnet, schirrmeister2017deep, saadati2019multimodal} and functional Near-Infrared Spectroscopy (fNIRS)~\cite{saadati2019multimodal, wang2021taking, mughal2021fnirs} data to measure brain activities. EEG and fNIRS data are spatio-temporal, in other words, these devices can capture data from multiple probes across the skull of participants, with specific layouts, and measure brain activity continuously in real-time.  

A review of the history of fNIRS is provided in~\cite{boas2014twenty}. fNIRS provides higher spatial resolution than EEG, making it possible to localize specific functional brain regions of activation, as could be done with the constrictive fMRI device~\cite{parasuraman2008neuroergonomics}. 
Furthermore, the new frequency-domain (FD)NIRS uses a linear symmetric dual-slope (DS) sensor design~\cite{blaney2020design}, which beneficially suppresses superficial hemodynamics, instrumental drifts and motion artifacts~\cite{blaney2020phase}.

Yet, the amount of publicly available fNIRS data has been limited, especially for studies covering a large number of participants. Available open access fNIRS datasets, such as TUBerlin~\cite{shin2018simultaneous}, usually contained data from 10-30 subjects. TUFTS~\cite{huang2021tufts} is a recently published open access fNIRS dataset, which has been collected from 68 participants. In addition to containing data from a larger number of participants and representing a larger variation, TUFTS datasets also provides a standard evaluation protocol for machine learning training paradigms. This makes training and testing different neural networks and comparing them more convenient and commensurate. For these reasons, we employ this newly published, larger scale TUFTS dataset to test our proposed method and compare it with different baselines.

\subsection{Workload classification with deep learning}
\label{subsec:deeplearning}

While traditional machine learning algorithms, such as Support Vector Machines (SVMs), Linear Discriminant Analysis (LDA), Principal Component Analysis (PCA), k-nearest neighbors algorithm (kNN), and Artificial Neural Networks (ANN), have been widely used for mental workload classification, deep learning-based algorithms have become more popular in recent years, especially thanks to their ability of extracting features and eliminating the need for handcrafted features. 

It has been reported that traditional machine learning algorithms, such as SVMs and LDA, provide good results for the detection of various levels of mental workload~\cite{aghajani2017measuring,hong2015classification,lee2014hybrid}. The main working principle of SVMs is finding a hyperplane that can separate the data accurately. If it results in a significantly smoother hyperplane in the optimization process, it provides a high generalization power by tolerating some misclassifications~\cite{benerradi2019exploring}. Compared to CNNs, SVMs are faster and affected less by random sets. However, SVMs employ handcrafted features and need selection of a best feature set, and their performance depends on the selection of these features and preprocessing. On the other hand, the main limitation of LDA in workload classification studies is its limited performance on nonlinear complex brain signals~\cite{garcia2003support}. 

Schirrmeister $\emph{et al.}$~\cite{schirrmeister2017deep} proposed a deep convolutional model, referred to as the DeepConv, for EEG signals, and showed that end-to-end deep CNNs trained within-subject can provide promising accuracy numbers. They first use 2D convolution (Conv2D) along temporal dimension followed by convolution over spatial dimension, and repeat this several times. At the end, they use a fully connected (FC) layer for final classification. EEGNet~\cite{lawhern2018eegnet} also shows promising results on EEG signals across different experiments covering visual-, memory- and movement-related scenarios. EEGNet uses depthwise separable convolution operators, which were introduced in the Xception architecture~\cite{chollet2017xception} for computer vision applications. EEGNet first uses Conv2D along temporal dimension to learn frequency filters, and then employs depthwise separable convolutions to combine frequency-specific features with spatial information for the final classifier, which is a single FC layer. Compared to DeepConvNet, EEGNet has far less trainable parameters. 
Saadati $\emph{et al.}$ \cite{saadati2019multimodal} proposed a CNN architecture for motor imagery and mental workload tasks, by using two types of brain data, namely fNIRS and EEG, together during training. The use of both modalities provides promising within-subject improvements compared to using single modality brain data. 

Besides CNNs, there are other types of neural network architectures commonly used in deep learning. Recurrent neural networks (RNNs) are widely used especially in application areas concerned with sequential or temporal data, such as text, audio and video~\cite{yu2019review}. Long Short Term Memory (LSTM) Networks~\cite{hochreiter1997long, graves2014neural} have been widely used to capture information along temporal dimension. 
Mughal $\emph{et al.}$~\cite{mughal2021fnirs} proposed a CNN-LSTM architecture for mental workload classification within subjects. 
Unlike EEGNet and DeepConvNet, this type of architecture first uses a CNN to encode the spatial information, and then employs LSTM to capture the temporal information. 

Aforementioned works mostly focus on within subject training. Cross-subject and cross-session CWL classification, on the other hand, is a much more challenging task due to the high variation in inter-subject fNIRS data and in intra-subject data across different sessions. Sommer $\emph{et al.}$~\cite{sommer2021classification} use CNN-LSTM on motor classification of finger tapping levels cross-sessions. Wang $\emph{et al.}$~\cite{wang2021taking} replace the LSTM with a gated recurrent unit (GRU)~\cite{chung2014empirical}, which has a similar structure to LSTM, while having less trainable parameters.  Wang $\emph{et al.}$~\cite{wang2021taking} also propose to use self-supervised label augmentation~\cite{lee2020self} by permuting the order of control rest-task pairs of samples as control rest-task pair and task-control rest pair. They present promising results on multi-label classification of working memory load (WML) and visual perceptual load (VPL) across different subjects. Lyu $\emph{et al.}$~\cite{lyu2021domain} view data from different sessions or from different subjects as data from different domains and use domain adaptation to align the data for n-back task classification. They show that this domain alignment works on SVM, CNN and RNN methods. 

The aforementioned methods employ CNNs to process spatial or temporal information. Yet, as discussed above, the assumption behind the parameter sharing aspect~\cite{alzubaidi2021review} of CNNs does not always hold for brain signals, 
since types and levels of workloads depend on specific brain areas~\cite{sabri2014neural}.

\section{Proposed Method}
\label{sec:method}
In this work, we focus on cross-subject Cognitive Workload (CWL) and Motion Workload (MWL) classification from fNIRS signals, and use data, collected during an n-back task, for evaluation. We propose an end-to-end MLPMixer classifier with a contrastive learning data sampler, where the contrastive samples are collected based on both across different subjects and different sessions. The proposed contrastive learning will only happened at training and won't adding extra calculations at implement.

\subsection{Motivation}
\label{subsec:impact_of_intra_variance}

we introduced how previous studies treat different subjects~\cite{zhong2020eeg} and sessions~\cite{lyu2021domain} as distinct domains and use domain adaptation (DA) to enable the extractor to create domain-invariant features, which can enhance the model’s generalization ability on unseen subjects or sessions. However, these previous works do not address the fact that intra-subject variance can also arise within the same session~\cite{huang2021tufts, yucel2021best}. We use a series of toy tasks to empirically show the distance and performance of different scenarios in terms of Wasserstein loss~\cite{arjovsky2017wasserstein} of the last layer of encoder and classification accuracy, respectively.  Wasserstein loss~\cite{arjovsky2017wasserstein} is widely used to measure the difference of two distributions. We compare the accuracy of two models, DeepConv~\cite{schirrmeister2017deep} (CNN-based model) and our MLPMixer (MLP-based model), on four different scenarios: split by (i) trail, (ii) block, (iii) session, and (iv) subject. For each scenario, we measure how well the models can classify the data from three-way n-back (0- vs. 1- vs. 2-back) on the TUBerlin dataset. These four scenarios are described in more detail as follows:
\begin{itemize}
\item[(i)] \textbf{Split by trail:} This scenario involves using all the data from all the subjects and sessions, for both training and testing and splitting by trails; 
\item[(ii)] \textbf{Split by block:} This scenario involves using all the data from all the subjects and sessions, for both training and testing and splitting by blocks; 
\item[(iii)] \textbf{Split by session:} This scenario involves using the data from two out of three session of each subject for training, and the data from the remaining session of each subject for testing; 
\item[(iv)] \textbf{Split by subjects:} This scenario involves using the data from some subjects for training, and the data from the remaining subjects for testing. 
\end{itemize}

\begin{figure}
    \centering
    \includegraphics[width=\linewidth]{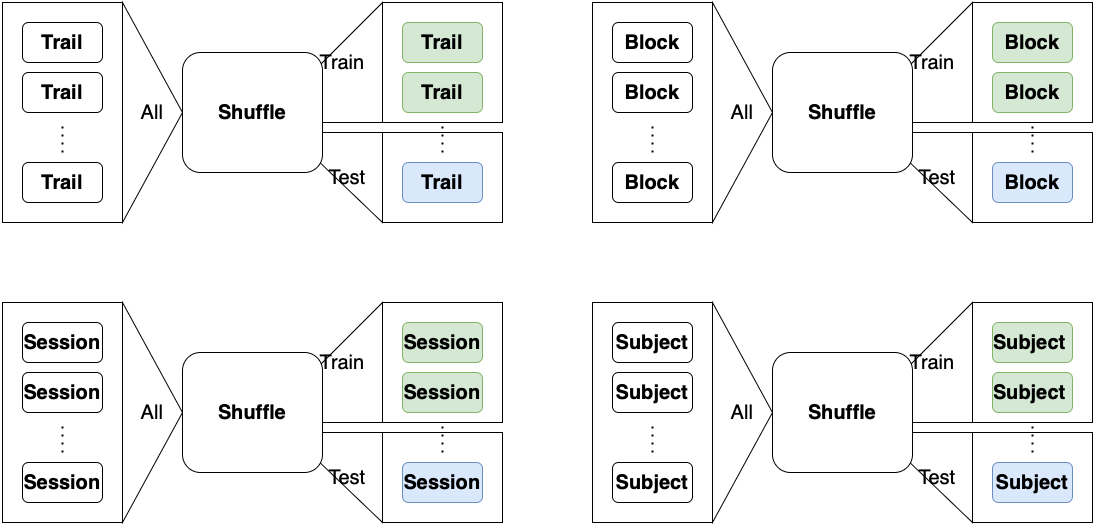}
    \caption{Four scenarios of splits.}
    \label{fig:splits_CBCA_motiv}
\end{figure}

\noindent Figure~\ref{fig:splits_CBCA_motiv} illustrates these four scenarios. If 
the total dataset is envisioned as a multidimensional tensor comprised of four axes (trial, block, session, and subjects), for each scenario, we execute the data split along the relevant axis, effectively dividing the dataset into distinct training and testing sets. This methodical partitioning ensures that our model training and evaluation processes are 
systematically structured allowing us to study effects of variations across trails, blocks, sessions and subjects. 

We use cross-entropy and train models supervised. 
Table~\ref{tab:ab1} summarizes the experimental results. As can be seen, the DeepConv model achieves an accuracy of $51.42\%$ when the data is split by trail. The accuracy decreases by $9.27\%$ when we change the data split format from trail to block, and the accuracy decreases by $9.07\%$ when we change the data split approach from trail to session. The accuracy drops by $8.62\%$ when we change the split approach from trail to subject. When the MLPMixer model is trained and tested on different splits of the data, its accuracy varies greatly. The highest accuracy of $87.53\%$ is achieved when the data is split by trail, meaning that the model sees data from all subjects, sessions, and blocks during training. However, when the data is split by session, the accuracy drops to $42.38\%$, indicating that the model struggles to generalize to new sessions of the same subjects. The accuracy further decreases to $40.44\%$ when the data is split by subject, showing that the model fails to transfer to new subjects. The lowest accuracy of $39.92\%$ is obtained when the data is split by block, suggesting that the model cannot handle the intra-variance between blocks of the same task or condition.

\begin{table}[h]
\centering
\begin{tabular}{|cc|c|c|c|c|}
\hline
\multicolumn{2}{|r|}{Split-by-}                       & Subject  & Session  & Block    & Trail             \\ \hline
\multicolumn{1}{|c|}{{DeepConv}} & WD  & 4.16E-03         & 5.65E-03         & 4.83E-03       & 3.20E-03 \\ \cline{2-6} 
\multicolumn{1}{|c|}{}                          & Acc & 42.80\%          & 42.35\%          & 42.15\%        & 51.42\%           \\ \hline
\multicolumn{1}{|c|}{{MLPMixer}} & WD  & 9.21E-03         & 9.31E-03         & 9.28E-03       & 5.97E-03          \\ \cline{2-6} 
\multicolumn{1}{|c|}{}                          & Acc & 40.44\%          & 42.38\%          & 39.92\%        & 87.53\%           \\ \hline
\end{tabular}
\caption{DeepConv and MLPMixer on TUberlin dataset for n-back classification task (0-, vs. 1- vs. 2-back) at different splitting scenarios. ``WD" and ``Acc" stand for Wasserstein distance and Accuracy, respectively.}
\label{tab:ab1}
\end{table}

In this work, DA-based approach is proposed to overcome the change of distribution between training and testing sets. We employ the Wasserstein distance as a metric to quantify the discrepancy between the features of training and testing sets under each scenario and report it as WD in Table~\ref{tab:ab1}. We measured the mean Wasserstein distance between different combinations of mini-batches from the training and testing data. We found that the lowest distance was achieved when the data was split by trail. This makes sense because the data in each trail was collected in a short and continuous time span. The other three ways of splitting the data (by block, by session and by subject) resulted in similar distances when we measure the features from the MLPMixer. The distance measurements from DeepConv show that the highest distance was obtained when the data was split by session, and splits by subject and by block resulted in similar distances. These sets of results show that the intra-subject difference between blocks should also be considered when developing new models for fNRIS data.

Our experiments above demonstrate that the models’ accuracy is highly affected by the intra-subject variance of the data within each block of the same task or condition. This variability is as influential as the variability of new sessions, as well as new subjects. This is a significant point that has not been addressed in previous studies~\cite{huang2021tufts, yucel2021best, lyu2021domain}. Motivated by this observation, we propose Class-Aware-Block-Aware Domain Adaptation (CABA-DA) to explicitly decrease the difference between blocks in the same session of the same subject.

\subsection{Preliminaries on Domain Adaptation-based Work}

Domain adaptation aims to better generalize and improve the performance of models on target domain. In general, it is assumed that the distribution of the source domain samples
$$\mathcal{S} = \{ (x_{1}^{s}, y_{1}^{s}), \cdots (x_{N_s}^{s}, y_{N_s}^{s})\}$$ 
\noindent and target domain samples
$$\mathcal{T} = \{ (x_{1}^{t}, y_{1}^{t}), \cdots (x_{N_t}^{t}, y_{N_t}^{t})\}$$ are different, where $(x_{i}^{s}, y_{i}^{s})$ is a pair of input $x_{i}^{s}$ and its label $y_{i}^{s}$ from the source domain, and $(x_{i}^{t}, y_{i}^{t})$ is a pair of input $x_{i}^{t}$ and its label $y_{i}^{t}$ from the target domain. 
To align the source and target domains, many domain adaptation methods use a feature extractor to create a representation that is invariant across domains. There are two common methods for achieving this. One is to minimize the distance between the features of the two distributions~\cite{sejdinovic2013equivalence, kang2019contrastive}. Another is to use a domain classifier in an adversarial way and reverse its gradient to make it fail to distinguish the domains~\cite{ajakan2014domain, zhong2020eeg}.
Treating data from different subjects as different domains and mitigating the domain shift via domain adaptation can improve the cross-subject performance. 
This idea has been been explored by Zhong et al.~\cite{zhong2020eeg} for brain-signal processing. To complicate matters further, in our problem setting, we also face with high intra-subject variance in data captured from the same subject during different sessions, and sometimes even during the same session. Hence, in this work, we propose to treat 

not only the data from different subjects but also different data blocks from the same subject as data from different domains. 

\subsection{Class Aware Block Aware Domain Adaptation}

In this section, we first review the work on class-aware domain adaption referred to as the contrastive domain discrepancy (CDD)~\cite{kang2019contrastive}. Then, we will introduce our class-aware and block-aware (CABA) domain adaptation method and sampling method. 

\noindent {\textbf{(i) Contrastive Domain Discrepancy (CDD)}
\par
CDD is developed from Maximum Mean Discrepancy (MMD)~\cite{sejdinovic2013equivalence}. In MMD, $\{x_i^{s}\}$ and $\{x_i^{t}\}$ are sampled from the marginal distributions $P(\mathbb{\mathrm{X^s}})$ and $Q(\mathbb{\mathrm{X^t}})$, respectively, which are independent and identically distributed (\textit{iid}). MMD is motivated by the fact that if two distributions are identical, all statistics should be the same. MMD uses the mean embeddings in the reproducing kernel Hilbert space (RKHS) to describe the difference between two distributions. In practice, for layer $l$ of a neural network, the squared value of MMD is estimated from the empirical kernel mean embeddings such that
\begin{equation}
    \begin{array}{rcl}
D_{l}^{mmd} & = & \frac{1}{n_{s}^{2}} \sum_{\substack{i=1}}^{\substack{n_s}} \sum_{\substack{j=1}}^{\substack{n_s}} k_l(\phi_l(x_i^s), \phi_l(x_j^s))  \\

 & + & \frac{1}{n_{t}^{2}} \sum_{\substack{i=1}}^{\substack{n_t}} \sum_{\substack{j=1}}^{\substack{n_t}} k_l(\phi_l(x_i^t), \phi_l(x_j^t)) \\

 & - & \frac{2}{n_{s}^{2}n_{t}^{2}} \sum_{\substack{i=1}}^{\substack{n_s}} \sum_{\substack{j=1}}^{\substack{n_t}} k_l(\phi_l(x_i^s), \phi_l(x_j^t)), 
\end{array}
\end{equation}

where $x^s\in\mathcal{S^{'}}\subset\mathcal{S}$, $x^t\in\mathcal{T^{'}}\subset\mathcal{T}$, and $\mathcal{S^{'}}$ and $\mathcal{T^{'}}$ are the mini-batch source and target data sampled from $\mathcal{S}$ and $\mathcal{T}$, respectively. $k_l$ denotes the kernel selected for the $l$-th layer of the neural network. 

CDD is established on MMD. It explicitly incorporates the class information into the formula and measures intra-class and inter-class discrepancy across domains. Minimizing the intra-class domain discrepancy can cluster representations of samples within the same class, whereas maximizing the inter-class domain discrepancy pushes the representations from different classes far away from each other. CDD defines class-aware domain discrepancy as 

\begin{equation}
    D_{c_{1}c_{2}}(\hat{y_{1}^{t}}, \hat{y_{2}^{t}} \cdots \hat{y_{n_t}^{t}}, \phi) = e_1 + e_2 - 2e_3
    \label{eq:domain-discrepancy}
\end{equation}

where 

\begin{equation}
    \begin{array}{rcl}
        e_1 & = & \sum_{\substack{i=1}}^{\substack{n_s}} \sum_{\substack{j=1}}^{\substack{n_s}} \frac{\mu_{c_1c_1} (y_i^s, y_j^s) k(\phi(x_i^s), \phi(x_j^s)) }{\sum_{i=1}^{n_s} \sum_{j=1}^{n_s}\mu_{c_1c_1}(y_i^s, y_j^s)} \\ 
        
        e_2 & = &  \sum_{\substack{i=1}}^{\substack{n_t}} \sum_{\substack{j=1}}^{\substack{n_t}} \frac{\mu_{c_2c_2} (y_i^t, y_j^t) k(\phi(x_i^t), \phi(x_j^t)) }{\sum_{i=1}^{n_t} \sum_{j=1}^{n_t}\mu_{c_2c_2}(y_i^t, y_j^t)} \\

        e_3 & = &  \sum_{\substack{i=1}}^{\substack{n_s}} \sum_{\substack{j=1}}^{\substack{n_t}} \frac{\mu_{c_1c_2} (y_i^s, y_j^t) k(\phi(x_i^s), \phi(x_j^t)) }{\sum_{i=1}^{n_s} \sum_{j=1}^{n_t}\mu_{c_1c_2}(y_i^s, y_j^t)} 
    \end{array}
    \label{eq:cdd}
\end{equation}

In Eq.~\ref{eq:domain-discrepancy}, $c1$ and $c2$ are classes to be used to calculate CCD. %
When $c1 = c2$, it measures intra-class domain discrepancy, when $c_1 \neq c_2$, it measures inter-class domain discrepancy. $\mu_{c_1c_2}$ is an indicator function such that
$$
     \mu_{cc^{'}}(y,y^{'}) = \left\{ \begin{array}{cl}
        1 & if y = c, y^{'} = c{'} \\
        0 & otherwise.
        \end{array} \right.
$$

Finally the CDD can be calculated as 

\begin{equation}
    \begin{array}{rcl}
        D^{cdd} & = &  \frac{1}{M}\sum_{\substack{c=1}}^{M} D_{cc}(y_{1:n_t}^{t}, \phi) \\
          & - & \frac{1}{M(M-1)} \sum_{\substack{c=1}}^{M} \sum_{\substack{c^{'}=1 \\ c'\neq c}}^{M}   D_{cc^{'}}(y_{1:n_t}^{t}, \phi)
    \end{array}
\end{equation}

Note that CDD~\cite{kang2019contrastive} was proposed for unsupervised domain adaptation, i.e. $y_{1:n_t}^{t}$ is unknown and needs to be estimated by network module. 

With fNIRS data, as will be discussed below,  we view samples from different blocks of subjects as different domains. In our case, $y_{1:n_t}^{t}$ stands for label of samples from one of the blocks of a subject in the training set. 

\noindent{\textbf{(ii) Proposed Class-Aware Block-Aware Domain Adaptation}}
\label{CABA}
\par
As shown in Tab.~\ref{tab:ab1}, the performance of the models is also significantly affected by the intra-subject variance across different blocks, which has a similar overall impact on the performance as the split by sessions and split by subjects. 
To address this issue, we propose to explicitly take the block information into account and measure the intra-class variance across blocks from the same session of subjects. 
Intuitively, brain signals that share the same label from the same subjects ought to exhibit similar representations. We propose leveraging the concept of inter-block discrepancy, by utilizing Maximum Mean Discrepancy (MMD) to evaluate the variations among blocks within the same session of a given subject. 

\begin{equation}
    \begin{array}{rcl}
        D^{caba}_{(z_1,z_2) | c} & = & \frac{1}{n_{z_1}^{2}} \sum_{\substack{i=1}}^{\substack{n_{z_1}}} \sum_{\substack{j=1}}^{\substack{n_{z_1}}} k_l(\phi_l(x_i^{z_1}), \phi_l(x_j^{z_1}))  \\
        
         & + & \frac{1}{n_{z_2}^{2}} \sum_{\substack{i=1}}^{\substack{n_{z_2}}} \sum_{\substack{j=1}}^{\substack{n_{z_2}}} k_l(\phi_l(x_i^{z_2}), \phi_l(x_j^{z_2})) \\
        
         & - & \frac{2}{n_{z_1}^{2}n_{z_2}^{2}} \sum_{\substack{i=1}}^{\substack{n_{z_1}}} \sum_{\substack{j=1}}^{\substack{n_{z_2}}} k_l(\phi_l(x_i^{z_1}), \phi_l(x_j^{z_2}))
    \end{array}
\end{equation}

where $z_1|c$ and $z_2|c$ are samples from different blocks in the same class $c$ of the same subject. Minimizing $D^{caba}_{(z_1,z_2) | c}$ can decrease the difference between two conditional distributions $P(X_{z_1}^s|Y^s)$ and $P(X_{z_2}^s|Y^s)$. In other words, the difference between the samples, with the same label from different blocks in the same session of the same subject, is minimized. 

\subsection{The Architecture of the Modified MLPMixer}
\label{subsec:mlpmixer}
MLPMixer~\cite{tolstikhin2021mlp} is fully built upon MLP layers. As mentioned above, the parameter sharing aspect of CNNs assumes that a patch of weights valid for one position is also useful for other locations, which does not always hold for fNIRS data. The reason is that  workload is related to brain areas, which are interconnected. On the other hand, the fully connected nature of MLPs is more suitable for analyzing fNIRS data.

MLPMixer~\cite{tolstikhin2021mlp} was proposed as an image classification model. It takes a sequence of non-overlapping image patches as input, and projects each patch to a desired hidden dimension as a token $F \in \mathbb{R}^{S \times C}$, where $S$ is the length of input sequence, and $C$ is the size of the hidden dimension of token. MLPMixer contains multiple Mixer layers, and each layer consists of two MLP blocks. The first MLP block is token-mixing MLP, which is applied to the columns of input (i.e. $F^\intercal$), and maps $\mathbb{R}^S  \mapsto \mathbb{R}^S $. The second MLP block is the channel-mixing MLP, which is applied to the rows of input $F$, and maps $\mathbb{R}^C \mapsto \mathbb{R}^C $. 
MLPMixer is a general backbone network for the image classification task. We modify the MLPMixer for the fNIRS classification task. 
The TUFTS dataset we use has data from two fNIRS channels (i.e. $\hat{x} \in \mathbb{R}^{2 \times T \times D}$). Thus, we add a fully connected layer ahead of MLPMixer, mapping $\mathbb{R}^{2\times D} \mapsto \mathbb{R}^{C}$, which corresponds to the token-mixing of the original MLPMixer. Figure~\ref{fig:mlpmixer_arch} shows the architecture of the MLPMixer we use. In our experiments, the parameters of the MLPMixer are set as follows: 
$T=150$, $D=4$, $C=16$, $N=4$. For the temporal-mixing MLP, the hidden dimension is 64, and for the channel-mixing MLP the hidden dimension is 32.

\begin{figure*}[h]
    \centering
    \includegraphics[width=\textwidth]{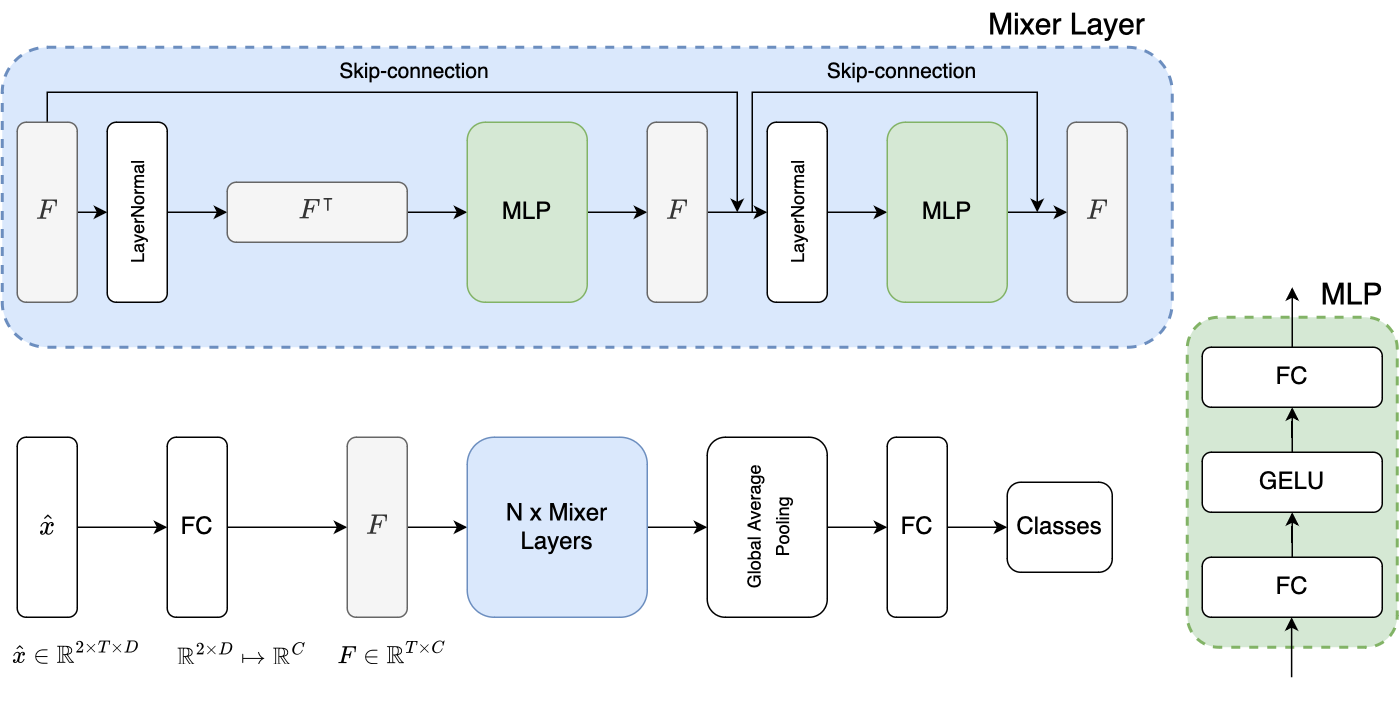}
    \caption{The MLPMixer-based classifier that we adapted for the fNIRS data. The input data $\hat{x} \in \mathbb{R}^{2 \times T \times D}$ is first encoded by an FC layer ($\mathbb{R}^{2\times D} \mapsto \mathbb{R}^{C}$). The remainder of this figure is the same as the original MLPMixer~\cite{tolstikhin2021mlp}: the encoded feature ($F$) is sent to $N$ mixer layers. Each mixer layer contains one temporal-mixing MLP ( $\mathbb{R}^T  \mapsto \mathbb{R}^T $) and one channel-mixing MLP ( $\mathbb{R}^C  \mapsto \mathbb{R}^C $). An MLP has two FC layers and a GELU nonlinearity. Other components include layer normalization, skip-connection and global average pooling.}
    \label{fig:mlpmixer_arch}
\end{figure*}

An MLP block can be written as follows:
\begin{equation}
    F = w_2\sigma(w_1(x) + b_1) + b_2,
\end{equation}
\noindent where $\sigma$ is an element-wise non-linearity GELU~\cite{hendrycks2016gaussian} layer and $w_1$ and $b_1$ are the weights and bias, respectively, for the first fully connected (FC) layer. They map the input feature to a hidden feature. More specifically, $\mathbb{R}^{T} \mapsto \mathbb{R}^{T_h}$ with temporal-mixing and $\mathbb{R}^{C} \mapsto \mathbb{R}^{C_h}$ with channel-mixing, where $T_h$ and $C_h$ are the hidden dimension for temporal-mixing and channel-mixing, respectively. $w_2$ and $b_2$ are the weights and bias, respectively, for the second FC layer mapping the hidden feature to the same dimension as the input feature. More specifically, $\mathbb{R}^{T_h} \mapsto \mathbb{R}^{T}$ with temporal-mixing and $\mathbb{R}^{C_h} \mapsto \mathbb{R}^{C}$ with channel-mixing. The output feature of an MLP block has the same dimension as the input feature to fit the residual skip-connection, which is a commonly used deep-learning approach proposed in ResNet~\cite{he2016deep}.

\subsection{The Overall Objective}

The overall objective can be written as

\begin{equation}
    \min_{\theta} l = l^{ce} + \alpha(D^{cdd} + D^{caba}),
\label{eq:obj}
\end{equation}

where $l^{ce}$ is the cross-entropy loss and $\alpha$ weights the discrepancy penalty term.  The minimization of the cross-entropy loss is aimed at supervised classification tasks. Minimizing $D^{cdd}$ aims to reduce the intra-class discrepancy while enhancing the inter-class discrepancy. Minimizing $D^{caba}$ is to minimize the intra-subjects difference. All three terms contribute to decreasing the performance drop, caused by domain shift and session inconsistency, when tested on brain signals of never-before-seen subjects.

\section{Experiments}
\label{sec:experiments}

In this section, we present the results of our experiments performed on three publicly available and commonly used fNIRS datasets. We first introduce these datasets. We then compare the performance of our proposed method with three different baselines, namely , DeepConv~\cite{schirrmeister2017deep}, EEGNet~\cite{lawhern2018eegnet} and MLPBiGRU~\cite{wang2021taking} on these datasets. Then, we present the visualization results. 

\subsection{Datasets and the Experiment Setup}
\label{sec:datasets}
We conduct our experiments on the TUberlin~\cite{shin2018simultaneous}, Tufts~\cite{huang2021tufts} and FingerFootTapping (FFT)~\cite{bak2019open} datasets.
TUberlin~\cite{shin2018simultaneous} and Tufts~\cite{huang2021tufts} datasets were collected from n-back tasks, and FignerFootTapping (FFT)~\cite{bak2019open} dataset was collected when participants were performing finger- and foot-tapping tasks.

\textbf{TUBerlin Dataset}: This dataset, proposed by Shin et al.,~\cite{shin2018simultaneous}, contains simultaneous EEG and fNIRS recordings of 26 participants, who performed three kinds of cognitive tasks: n-back, discrimination/selection response task (DSR), and word generation (WG) tasks. The dataset is open-access and multimodal, capturing brain activity from different sources. 
We use the data from the n-back task to test our method. The n-back task involves 0-, 2-, and 3-back levels, and our aim is to classify the mental workload levels across subjects according to the difficulty of n-back tasks. As $n$ increases, the working memory workload increases.
The dataset of the n-back task consists of three sessions. In each session, the subjects completed three series of 0-, 2-, and 3-back tasks in a counterbalanced order (i.e. 0 $\to$ 2 $\to$ 3 $\to$ 2 $\to$ 3 $\to$ 0 $\to$ 3 $\to$ 0 $\to$ 2). There are nine series of n-back tasks in total for each subject. Each blocks starts with a 2-second instruction that shows the task type (0-, 2-, or 3-back), followed by a 40-second task period, and ends with a 20-second rest period. 
During the task period, a single-digit number was randomly presented every 2 seconds, resulting in twenty trails per block. For each n-back task, there are 180 trails in total ( = 20 trails $\times$ 3 blocks $\times$ 3 sessions).

fNIRS data is acquired by NIRScout device (NIRx Medizintechnik GmbH, Berlin, Germany) and further converted to deoxy-(HbR) and oxy hemoglobin (HbO) intensity changes, using modified Beer-Lambert law~\cite{kocsis2006modified}, and downsampled to 10Hz. Then, the downsampled data was low- pass filtered instead of band-pass, (6th order zero-phase Butterworth) with 0.2 Hz cut-off frequency to remove the high frequency instrument and systemic physiological noise.
It has sixteen sources and sixteen detectors that were positioned on the frontal, motor, parietal, and occipital regions of the head. An fNIRS channel is formed by a source-detector pair that was next to each other resulting 36 channels in total. Each channel has 2 features corresponding to the $\Delta$HbR and $\Delta$HbO data. The total number of features is 72, i.e. 36 spatial locations  $\times$ 2 hemoglobin types $\times$ 1 optical data type (intensity). 

\textbf{FFT Dataset}: Bak et al.~\cite{bak2019open} proposed an open-access fNIRS dataset, which has data from 30 participants for 
three-class classification, namely left-hand unilateral complex finger-tapping (LHT) (Class 0) , right-hand unilateral complex finger-tapping (RHT) (Class 1) , and foot-tapping (FT) (Class 2) .
In each session, the order of task is randomly generated. A trail starts with a 2-seconds of introduction and a 10 seconds of task period followed by an 17-19 seconds inter-trail break. There are total of 225 trails (= 25 trails $\times$ 3 task types $\times$ 3 sessions). 

fNIRS data were recorded by a three-wavelength continuous-time multi-channel fNIRS system (LIGHTNIRS, Shimadzu, Kyoto, Japan) consisting of eight sources and eight detectors. Four of the sources and detectors  were placed around C3 on the left hemisphere, and the rest were placed around C4 on the right hemisphere. The raw fNRIS data was further converted to the intensity changes i.e. $\Delta$HbR and $\Delta$HbO using modified Beer-Lambert law~\cite{kocsis2006modified} with sample rate at 13.33Hz. Then, data was band-pass filtered through a zero-order filter implemented by the third-order Butterworth filter with a pass-band of 0.01–0.1 Hz to remove the physiological noises. 
It contains 20 fNIRS channels and the total number of features is 40, i.e. 20 spatial locations (10 for each hemisphere) $\times$ 2 hemoglobin types $\times$ 1 optical data type (intensity).

\textbf{Tufts Dataset}: Recently, Huang et al.~\cite{huang2021tufts} proposed the largest open-access fNIRS dataset including data from 68 participants performing n-back task. The n-back task involves 0-, 1-, 2- and 3-back levels, and our aim is binary classification between 0- and 2- levels, which is same as the baselines in Huang et al.~\cite{huang2021tufts}. Each subject has only one session, and completed 16 blocks of n-back trails in a counterbalanced order (i.e. 0 $\to$ 1 $\to$ 2 $\to$ 3 $\to$ 1 $\to$ 2 $\to$ 3 $\to$ 0 $\to$ 2 $\to$ 3 $\to$ 0 $\to$ 1 $\to$ 3 $\to$ 0 $\to$ 1 $\to$ 2). Each block contains 40 trails lasting a total of 80 seconds (each trail lasting 2 seconds), followed by 10-30 seconds of rest period. %
For each participant, there are 640 trails in total (= 40 trails $\times$ 16 blocks $\times$ 1 session). 

fNIRS data is acquired by an Imagent frequency-domain (FD) NIRS instrument manufactured by ISS (Champaign, IL, USA). Two sets (left and right) of custom probes with linear symmetric dual-slope (DS)~\cite{blaney2020design} sensor design were placed at forehead. The raw data was further converted to the changes of HbR and HbO in intensity and phase~\cite{blaney2020phase} and sampled at 10Hz. Then, each univariate time-series was bandpass filtered using a 3rd-order zero-phase Butterworth filter, retaining 0.001-0.2 Hz to remove noise.
The data has total of 8 features, i.e. 2 spatial locations $\times$ 2 hemoglobin types $\times$ 2 optical data types (intensity and phase). 

\textbf{Experiment Setup}: We used the data from all three datasets as it is, without further preprocessing, since data had already been filtered with band-pass/low-pass filters to eliminate noise. We created the input data with sliding windows. The window size was the same as the task period of each trail, i.e. 2 seconds for TUBerlin and 10 seconds for FFT. For Tufts dataset, we used a 15-second window as recommended by the original paper. 
The input shapes and corresponding sliding window duration for all three datasets we used in our experiments are listed in Table \ref{tab:data_summary}. It is worth to mention that, in FFT, the experiments are not in block-wise design. Thus, for FFT, we applied our proposed CABA-DA on sessions instead of blocks.

\begin{table}[h]
\centering
\begin{tabular}{|c|c|c|}
\hline
Dataset  & Input Shape & Duration \\ \hline
TUberlin & $x \in \mathbb{R}^{20\times72}$           & 2 s      \\ \hline
FFT      & $x \in \mathbb{R}^{134\times40}$          & 10 s     \\ \hline
Tufts    & $x \in \mathbb{R}^{150\times8}$           & 15 s     \\ \hline
\end{tabular}
\caption{Input properties: Input sample $x \in \mathbb{R}^{S\times D}$, where $S$ and $D$ represent sequence length and number of features, respectively.}
\label{tab:data_summary}
\end{table}

All models, including our proposed method and the baseline models, were trained from scratch. Training was stopped when the evaluation loss did not improve for 50 consecutive epochs. We used the Adam optimizer~\cite{kingma2014adam} for training. We performed a grid search to find the best hyper-parameters, namely the learning rate and the dropout ratio for all models. The learning rate was chosen from \{$1e^{-4},1e^{-3},1e^{-2},1e^{-1}$\} and the dropout ratio was chosen from \{$0, 0.25, 0.5, 0.75$\}. The value of $\alpha$ in Eq.~\ref{eq:obj} is selected by grid search from \{0.5, 0.6, 0.7, 0.8, 0.9, 1.0, 1.1\} based on the test mean accuracy of the MLPMixer model. $\alpha$ was set as 1.1 for Tufts and, and as 1.0 for TUberlin and FFT datasets.

\subsection{Experimental Results}
We have conducted four sets of experiments on three open-access datasets and compared our MLPMixer model with three baselines, namely DeepConv, EEGNet and MLPBiGRU. We used $k$-fold cross-subject validation and reported the mean accuracy over $k$ folds. For TUBerlin and FFT datasets, we divided the data by subject IDs into 10 folds, with each fold containing data from only one participant. For the Tufts dataset, we used the original splits provided in the paper~\cite{huang2021tufts}, which also separated the folds by participants. Experimental results obtained on TUBerlin, FFT and Tufts datasets are shown in Tables~\ref{tab:res-Tuberlin},~\ref{tab:res-fft} and~\ref{tab:res-Tufts}, respectively. In our presentation of results, we highlight the highest value in each column in bold, and the highest value in the entire table is both bolded and underlined for clear distinction. For domain adaptation (DA) methods, the differences between the DA outcomes and those obtained through training with cross-entropy (CE) are displayed in parentheses. A green color indicates an increase in performance due to the DA method, while a red color signifies a decrease.

In the first set of experiments, we only used cross-entropy (CE in tables), without domain adaptation, to train proposed MLPMixer and the three baselines. Tables~\ref{tab:res-Tuberlin},~\ref{tab:res-fft} and~\ref{tab:res-Tufts} show that our proposed MLPMixer achieved a mean accuracy of $41.47\%$ on the three-way workload n-back (0- vs. 2- vs. 3-back) classification task on TUBerlin dataset, which is $1.56\%$ lower than the best accuracy obtained by EEGNet. However, our MLPMixer provided $64.76\%$ accuracy on the three-way LHT vs. RHT vs. FT classification task on the FFT dataset, outperforming all three baselines. 
For the binary workload n-back (0- vs. 2-back) classification task on the Tufts dataset, our MLPMixer again provided the highest mean accuracy of $67.28\%$, which is $0.95\%$ higher that the second best accuracy obtained by MLPBiGRU. From these results, we can see that our proposed MLPMixer model shows better generalizability in handling fNIRS data obtained in different settings and for different tasks across subjects.
\begin{table*}[h]
\resizebox{\linewidth}{!}{
\begin{tabular}{|c|c|c|c|c|}
\hline
Model    & CE(no DA) & Subject DA     & Session DA  & CABA-DA (Blockwise DA)         \\ \hline
DeepConv~\cite{schirrmeister2017deep} & 42.67\%   & \textbf{42.59\%} (\textcolor{BrickRed}{-0.08\%})          & \textbf{42.69\%} (\textcolor{ForestGreen}{0.02\%})   & 43.00\% (\textcolor{ForestGreen}{0.33\%}) \\ \hline
EEGNet~\cite{lawhern2018eegnet}       & \textbf{43.03\%}   & 42.36\% (\textcolor{BrickRed}{-0.67\%})         & 41.70\% (\textcolor{BrickRed}{-1.33\%})   & 42.44\% (\textcolor{BrickRed}{-0.59\%}) \\ \hline
MLPBiGRU~\cite{wang2021taking}        & 42.77\%   & 41.10\% (\textcolor{BrickRed}{-1.67\%})         & 42.63\% (\textcolor{BrickRed}{-0.14\%})   & \underline{\textbf{43.35\%}}(\textcolor{ForestGreen}{0.58\%}) \\ \hline
MLPMixer (ours)                       & 41.47\%   & 41.53\% (\textcolor{ForestGreen}{0.06\%})         & 42.28\% (\textcolor{ForestGreen}{0.81\%})   & 42.78\% (\textcolor{ForestGreen}{\textbf{1.31\%}})\\ \hline
\end{tabular}
}
\caption{Three-way workload n-back (0-, vs. 2- vs. 3-back) task classification results on TUBerlin dataset. We report mean accuracy over 10 folds without DA (CE) and with DA by using Subject, Session and Block as different domains at each experiment.}
\label{tab:res-Tuberlin}
\end{table*}

\begin{table*}[h!]
\resizebox{\linewidth}{!}{
\begin{tabular}{|c|c|c|c|c|}
\hline
Model                                   & CE(no DA)        & Subject DA & Session DA & CABA-DA (Blockwise DA)         \\ \hline
DeepConv~\cite{schirrmeister2017deep}   & 64.42\%          & 63.11\% (\textcolor{BrickRed}{-1.31\%})              & 62.89\% (\textcolor{BrickRed}{-1.53\%})                 & 64.64\% (\textcolor{ForestGreen}{0.22\%}) \\ \hline
EEGNet~\cite{lawhern2018eegnet}         & 60.71\%          & 60.76\% (\textcolor{ForestGreen}{0.04\%})            & 60.93\% (\textcolor{ForestGreen}{0.22\%})               & 60.99\% (\textcolor{ForestGreen}{0.28\%}) \\ \hline
MLPBiGRU~\cite{wang2021taking}          & 60.62\%          & 62.98\% (\textcolor{ForestGreen}{2.36\%})            & 62.76\% (\textcolor{ForestGreen}{2.13\%})               & 63.38\% (\textcolor{ForestGreen}{\textbf{2.76\%}}) \\ \hline
MLPMixer (ours)                         & \textbf{64.76\%}          & \textbf{64.58\%} (\textcolor{BrickRed}{-0.18\%})     & \textbf{65.73\%} (\textcolor{ForestGreen}{0.98\%})      & \underline{\textbf{65.87\%}}(\textcolor{ForestGreen}{1.11\%}) \\ \hline
\end{tabular}
}
\caption{LHT vs. RHT vs FT classification results on FFT. We report mean accuracy over 10 folds without DA (CE) and with DA but view Subject, Session and Block as different domains. FFT is not collected in block design, but we view the consecutive 3 trails as one block.}
\label{tab:res-fft}
\end{table*}

\begin{table*}[h]
\resizebox{\linewidth}{!}{
\begin{tabular}{|c|r|r|c|c|}
\hline
Model    & \multicolumn{1}{c|}{CE(no DA)} & \multicolumn{1}{c|}{Subject DA} & Session DA & \multicolumn{1}{c|}{CABA-DA (Blockwise DA)} \\ \hline
DeepConv~\cite{schirrmeister2017deep}   & 63.75\%                        & 61.96\% (\textcolor{BrickRed}{-1.79\%})            & -               & 67.56\% (\textcolor{ForestGreen}{3.80\%})             \\ \hline
EEGNet~\cite{lawhern2018eegnet}         & 62.08\%                        & 65.37\% (\textcolor{ForestGreen}{3.29\%})   & -             & 66.51\% (\textcolor{ForestGreen}{\textbf{4.43\%}})             \\ \hline
MLPBiGRU~\cite{wang2021taking}          & 66.33\%                        & 67.26\% (\textcolor{ForestGreen}{0.92\%})            & -             & 67.40\% (\textcolor{ForestGreen}{1.07\%})             \\ \hline
MLPMixer (ours)                         & \textbf{67.28\%}               & \textbf{67.72\%} (\textcolor{BrickRed}{0.45\%})              & -              & \underline{\textbf{67.91\%}} (\textcolor{ForestGreen}{0.63\%})             \\ \hline
\end{tabular}
}
\caption{Binary workload n-back (0- vs. 2-back) task classification results on TUFTS generic (cross subject) paradigm. We report mean accuracy over 17 folds without DA (CE) and with DA but view Subject, Session and Block as different domains. Tufts dataset has only one session.}
\label{tab:res-Tufts}
\end{table*}

To assess the impact of CABA-DA on three baseline models and our proposed MLPMixer model, we performed a second set of experiments. 
The results on the TUBerlin dataset (last column of Table~\ref{tab:res-Tuberlin}) show that our proposed CABA-DA approach improves the performance of all models except EEGNet. The most significant improvement was achieved by applying CABA to the MLPMixer model (which was proposed in Chapter 3), which increased the accuracy from $41.47\%$ to $42.78\%$, a gain of $1.31\%$. As for the FFT dataset (Table \ref{tab:res-fft}), our proposed CABA-DA approach also improved the performance of all models. The MLPBiGRU model experienced the largest improvement by using CABA-DA, which raised the accuracy from $60.62\%$ to $63.38\%$, an improvement of $2.76\%$. 
All models performed better on the Tufts dataset after applying our proposed CABA-DA. The EEGNet model benefited the most from CABA-DA, as its accuracy was boosted from $62.08\%$ to $66.51\%$, an increase of $4.43\%$.

To explore the effect of treating samples from different blocks as different domains versus treating different subjects and sessions as different domains, we conducted two additional sets of experiments. When we treat different subjects as different domains, the second term of Eq.~(\ref{eq:obj}), i.e. $D^{caba}$, reduces to zero. When we treat different sessions as different domains, then $D^{caba}$ measures the discrepancy between samples with the same label from different sessions of the same subject. The results are presented in the third and fourth columns of Tables~\ref{tab:res-Tuberlin}, \ref{tab:res-fft} and~\ref{tab:res-Tufts}. Since there was only one session for each subject in the TUFTS dataset, the Session DA column of Table \ref{tab:res-Tufts} is not applicable. In all 12 cases, CABA-DA (with block wise DA) outperforms Subject-DA and Session-DA on the TUBerlin and FFT datasets. For the TUFTS dataset, CABA-DA outperforms Subject-DA for all models. As for the comparison of Subject-DA versus Session-DA, for the DeepConv model, Session-DA outperformed Subject-DA on the TUBerlin dataset. Using EEGNet, Session-DA also achieved better results than Subject-DA on the FFT dataset. 
Overall, 11 out of 12 cases, CABA-DA outperformed CE (not using domain adaptation). And all the best performances are reached by using CABA-DA for all three datasets.

Our results demonstrate that our MLPMixer model has a high generalizability on fNIRS data for different kinds of tasks, such as three-way/binary n-back classification and LHT vs. RHT vs. FT motion workload classification task. We also show that our proposed CABA-DA method is versatile and effective when used with different kinds of networks, such as CNN-based (DeepConv and EEGNet), RNN-based (MLPBiGRU) and MLP-based (MLPMixer) methods, for the cognitive workload classification task. Moreover, we observe similar improvements on motion workload classification task.  

\begin{table}[h]
\centering
\begin{tabular}{|l|l|l|l|}
\hline
MACs                            & TUBerlin & FFT     & TUFTS  \\ \hline
DeepConv
~\cite{schirrmeister2017deep}   & 956.3K   & 6.03M   & 3.12M  \\ \hline
EEGNet~\cite{lawhern2018eegnet} & 30.47K   & 118.22K & 35.36K \\ \hline
MLPBiGRU~\cite{wang2021taking}  & 904.71K  & 5.94M   & 6.58M  \\ \hline
MLPMixer(ours)                  & 409.76K  & 2.4M    & 2.37M  \\ \hline
\end{tabular}
\caption{Comparing computation cost at inference.}
\label{tab:macs_report}
\end{table}

Our experiments across three public fNIRS datasets demonstrate that the proposed CABA-DA method can generally enhance performance across various classifier types, including those based on CNNs, RNNs, and MLPs. In terms of computational complexity, our method does not add extra modules to networks. However, compared to conventional training with cross-entropy, training with CABA-DA incurs additional computational effort due to the calculation of CDD and CABA, necessitating the pairing of samples for network training. Given the shallow and lightweight nature of these networks, the increase in computational demand is minimal. Modern computers equipped with graphics cards can easily handle this additional computational cost. Importantly, at the inference stage, CABA-DA introduces no extra computational expenses.
In our comparison of the Multiply-Accumulate operations (MACs) between baseline models and our proposed MLPMixer model across three datasets at inference, as detailed in Table~\ref{tab:macs_report}, we observed that EEGNet requires the least computational cost. Our MLPMixer model ranks as the second least computationally intensive among the evaluated models. 
With the implementation of CABA-DA, EEGNet achieved the highest accuracy on the TUBerlin and FFT datasets, whereas our MLPMixer model attained the highest accuracy on the TUFTS dataset. The compactness of EEGNet, being the smallest model evaluated, significantly minimizes the risk of overfitting, a crucial advantage for smaller datasets. Specifically, the TUBerlin and FFT datasets include 22 and 30 participants, respectively, compared to the larger TUFTS dataset, which comprises 68 participants. This compactness, while beneficial in preventing overfitting on smaller datasets, may limit EEGNet's performance potential in scenarios with more samples that could train larger models. 

\subsection{Visualization}
\label{subsec:visual}
In this section, we visualize the brain regions that are most influential by measuring how the performance changes when some fNIRS channels are masked, following a similar approach to what we described in Chapter 3. To determine the most important areas for brain-computer interface (BCI) performance, we used two datasets with high-density fNIRS channels, namely TUBerlin and FFT. We did not use the Tufts dataset, which only had two probes on the frontal cortex.

\begin{figure}[h]
    \centering
    \subfloat[]{\includegraphics[width=0.5\linewidth]{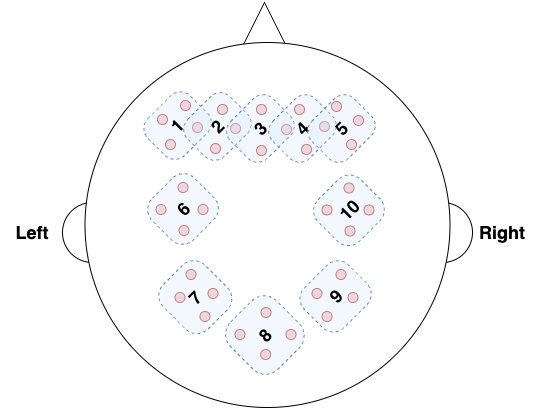}}
    ~
    \subfloat[]{\includegraphics[width=0.5\linewidth]{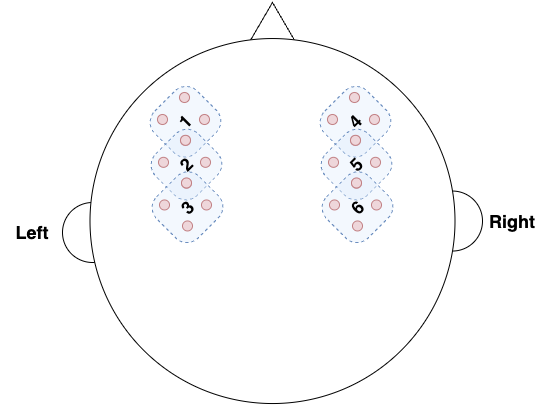}}
    \caption{The fNIRS channels and masks for the (a) TUBerlin and (b) FFT dataset. The red dots represent the fNIRS channels and the masks are numbered. There are 36 channels and 10 masks for the TUBerlin dataset, and 20 channels and 6 masks for the FFT dataset. Each mask covers 4 channels.}
    \label{fig:visual_diagram}
\end{figure}

Figure ~\ref{fig:visual_diagram} shows the fNIRS channels of the TUBerlin and FFT datasets with the numbered masks. For TUBerlin dataset, we selected the MLPBiGRU-Block-DA model, which achieved the best accuracy and out of the same reaso, we selected MLPMixer-Block-DA on FFT, for further analysis. We applied each mask one at a time and measured the change in accuracy of the trained model. More specifically, at each mask, we set corresponding values to 0 to block out the information contained at that position. 
Figure~\ref{fig:performance_masked} shows the results of this experiment. The black dotted lines indicate the average accuracy over all channels after masking for each dataset. The channels that caused a significant drop in accuracy, below the lower bound of the 95\% confidence interval (green dotted line), were considered critical channels. 

For the TUBerlin dataset, the critical channels were 2, 7 and 9, which correspond to the AF3, P3 and P4 areas~\cite{shin2018simultaneous, oostenveld2001five}, respectively. Where AF3 area is located in the left of the midline of the prefrontal cortex which are considered as a part of workload memory. P3 area is located in the left parietal lobe.  Mirroring P3, P4 is located over the right parietal lobe. Our experiments show that the prefrontal cortex and parietal lobes will work together to solve the n-back task. For the FFT dataset, mask position 2 and 5 have lower accuracy than the 95\% CI lower bound, which are associated with C3 and C4, respectively~\cite{bak2019open}, which are situated within the primary motor cortex. Our findings are consistent with this and also show that the C3 and C4 regions are predominantly associated with the processing of motion-related workload.

\begin{figure}[H]
    \centering
    \subfloat[]{\includegraphics[width=0.5\linewidth]{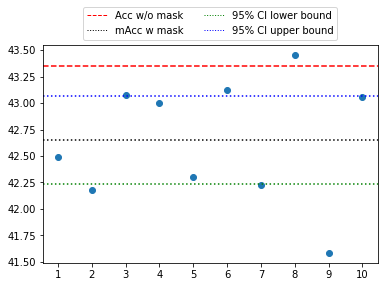}}
    ~
    \subfloat[]{\includegraphics[width=0.5\linewidth]{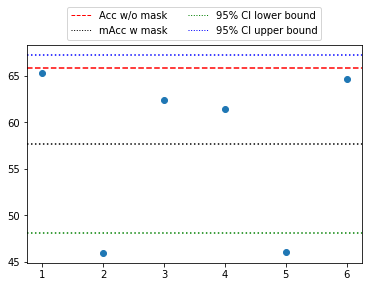}}
    \caption{Plot of of masking one position at a time using  (a) MLPBiGRU-Block-DA on TUBerlin and (b) MLPMixer-Block-DA on FFT. The x-axis shows the position of the mask and the y-axis shows the k-fold cross validation accuracy on test splits. The black dotted line represents the average performance across all mask positions. The red dash line shows the performance without any mask. The green and blue dot-dash lines indicate the lower and upper bounds of the 95\% confidence intervals of the masked accuracy.}
    \label{fig:performance_masked}
\end{figure}

\section{Ablation Studies}
\subsection{Effectiveness of Block Discrepancy Term}
We have conducted new variants of our main experiments using the same training settings. Our objective function in Eq.~(\ref{eq:obj}) consists of two terms: the first term is a cross entropy loss that provides the supervised signal for classification, and the second term is a combination of two discrepancy terms: $D^{cdd}$ and $D^{caba}$. The term $D^{cdd}$ is for contrastive learning, and it minimizes the intra-class discrepancy and maximizes the inter-class discrepancy. The term $D^{caba}$ is introduced by our approach, and minimizes the intra-subject difference across blocks. To evaluate the effectiveness of our proposed $D^{caba}$ term, we removed it from the objective function and wrote the new objection function as follows:
\begin{equation}
    \min_{\theta} l = l^{ce} + \alpha D^{cdd}.
    \label{eq:ab_obj}
\end{equation}

Then, we performed the experiments with this new objective function and our proposed MLPMixer on the TUBerlin, FFT and Tufts datasets. The results presented in Table~\ref{fig:ab_B} show that the block discrepancy term is indeed effective for improving the accuracy of the MLPMixer model. When this term is eliminated, the accuracy of the MLPMixer model on TUBerlin, FFT and Tufts datasets decreases by $0.47\%$, $2.14\%$ and $0.77\%$, respectively. This indicates that the block discrepancy term helps the model to get a better performance. The reason for this is that $D^{caba}$ is a term that explicitly measures the difference of the distributions of different blocks within the same session. By minimizing this term, we reduce the intra-subject variance across blocks. 
\begin{table}[h]
\centering
\begin{tabular}{|c|c|c|}
\hline
Dataset  & without $D^{caba}$ & with $D^{caba}$ \\ \hline
TUberlin & 42.31\% & \textbf{42.78\%} \\ \hline
FFT      & 63.73\% & \textbf{65.87\%} \\ \hline
Tufts    & 67.14\% & \textbf{67.91\%} \\ \hline
\end{tabular}
\caption{Accuracy of MLPMixer trained with and without block discrepancy term ($D^{caba}$).}
\label{fig:ab_B}
\end{table}

\subsection{Role of $\alpha$}

In this section, we train our model using CABA-DA and with different values of $\alpha$ (Eq.~(\ref{eq:obj})). As seen in Eq.~(\ref{eq:obj}), $\alpha$ weights the $D^{caba}$ and $D^{cdd}$ terms in the objective function. We conduct our experiments on Tufts dataset, since it has the largest amount of participants and the original paper~\cite{huang2021tufts} has released 17-folds splitting. We report the grid search of alpha starting from 0.0 to 1.1 with step size of 0.1. The mean accuracy of our MLPMixer is reported in Fig.~\ref{fig:effectiveness of CABA}.
The plot shows how the mean accuracy of MLPMixer varies with different values of $\alpha$.
The mean accuracy is calculated by averaging the accuracy of 17-fold cross-validation across participants. 
The plot reveals that there is in general a positive relationship between $\alpha$ and mean accuracy. 
The highest mean accuracy of $67.91\%$ is achieved when $\alpha$ is set to 1.0. 
\begin{figure}[h]
    \centering
    \includegraphics[width=\linewidth]{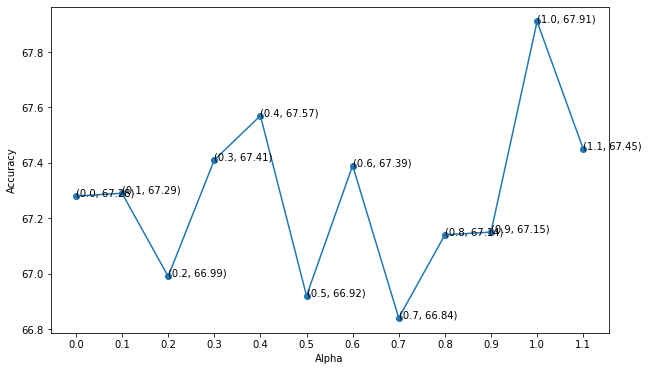}
    \caption{Plot of the mean accuracy (y-axis) of our MLPMixer for 17-fold cross-validation across participants on Tufts dataset against different values of $\alpha$ on the x-axis.}
    \label{fig:effectiveness of CABA}
\end{figure}

\section{Summary}
\label{sec:summ}
ne of the major challenges with CWL classification in real-world scenarios is the high variability of fNIRS data across different subjects and sessions, which demands a model that can generalize well to never-before-seen participants. Most of the existing works have adopted a per-participant approach for model training. Some more recent works have employed domain adaptation (DA)  to reduce the high variability by treating different subjects or sessions as different domains. We have demonstrated that the variability for the same subject across different blocks is also significant, by comparing the accuracy of different splitting methods for the same dataset. We have also computed the Wassertein distance to measure the level of difference between training and testing with different splitting methods, and found that splitting by blocks has similar level of difference as splitting by subjects. 
To alleviate the influence of intra-subject variance and improve the model generalizability, we have proposed to view blocks as different domains. To achieve this, we have proposed a Class-Aware Block-Aware (CABA) term to explicitly measure the difference of the distributions of different blocks within the same session of the same subject. 
We have also discussed the limitation of the CNN models that assume spatial invariance, which is not suitable for fNRIS data. We have proposed to use an MLPMixer-based approach instead of CNNs. Our experimental results have shown that MLPMixer as the baseline model (trained only with cross entropy) has achieved a significant improvement on FFT and Tufts datasets. Our proposed CABA-DA has consistently enhanced the performance of three out of four models on TUBerlin and all four models on Tufts dataset for n-back task, and has improved the performance of all four moels on the FFT dataset for finger and foot tapping task. 
\bibliographystyle{ieeetr}
\bibliography{refs}

\begin{thebibliography}{10}

\bibitem{mandrick2016neural}
K.~Mandrick, V.~Peysakhovich, F.~R{\'e}my, E.~Lepron, and M.~Causse, ``Neural
  and psychophysiological correlates of human performance under stress and high
  mental workload,'' {\em Biological psychology}, vol.~121, pp.~62--73, 2016.

\bibitem{dimoka2012conduct}
A.~Dimoka, ``How to conduct a functional magnetic resonance (fmri) study in
  social science research,'' {\em MIS quarterly}, pp.~811--840, 2012.

\bibitem{chance1993cognition}
B.~Chance, Z.~Zhuang, C.~UnAh, C.~Alter, and L.~Lipton, ``Cognition-activated
  low-frequency modulation of light absorption in human brain.,'' {\em
  Proceedings of the National Academy of Sciences}, vol.~90, no.~8,
  pp.~3770--3774, 1993.

\bibitem{shin2018simultaneous}
J.~Shin, A.~Von~L{\"u}hmann, D.-W. Kim, J.~Mehnert, H.-J. Hwang, and K.-R.
  M{\"u}ller, ``Simultaneous acquisition of eeg and nirs during cognitive tasks
  for an open access dataset,'' {\em Scientific data}, vol.~5, no.~1,
  pp.~1--16, 2018.

\bibitem{saadati2021application}
M.~Saadati, J.~Nelson, A.~Curtin, L.~Wang, and H.~Ayaz, ``Application of
  recurrent convolutional neural networks for mental workload assessment using
  functional near-infrared spectroscopy,'' in {\em International Conference on
  Applied Human Factors and Ergonomics}, pp.~106--113, Springer, 2021.

\bibitem{saadati2019mental}
M.~Saadati, J.~Nelson, and H.~Ayaz, ``Mental workload classification from
  spatial representation of fnirs recordings using convolutional neural
  networks,'' in {\em 2019 IEEE 29th International Workshop on Machine Learning
  for Signal Processing (MLSP)}, pp.~1--6, IEEE, 2019.

\bibitem{saadati2019convolutional}
M.~Saadati, J.~Nelson, and H.~Ayaz, ``Convolutional neural network for hybrid
  fnirs-eeg mental workload classification,'' in {\em International Conference
  on Applied Human Factors and Ergonomics}, pp.~221--232, Springer, 2019.

\bibitem{saadati2019multimodal}
M.~Saadati, J.~Nelson, and H.~Ayaz, ``Multimodal fnirs-eeg classification using
  deep learning algorithms for brain-computer interfaces purposes,'' in {\em
  International Conference on Applied Human Factors and Ergonomics},
  pp.~209--220, Springer, 2019.

\bibitem{mughal2021fnirs}
N.~E. Mughal, K.~Khalil, and M.~J. Khan, ``fnirs based multi-class mental
  workload classification using recurrence plots and cnn-lstm,'' in {\em 2021
  International Conference on Artificial Intelligence and Mechatronics Systems
  (AIMS)}, pp.~1--6, IEEE, 2021.

\bibitem{asgher2020classification}
U.~Asgher, K.~Khalil, Y.~Ayaz, R.~Ahmad, and M.~J. Khan, ``Classification of
  mental workload (mwl) using support vector machines (svm) and convolutional
  neural networks (cnn),'' in {\em 2020 3rd International Conference on
  Computing, Mathematics and Engineering Technologies (iCoMET)}, pp.~1--6,
  IEEE, 2020.

\bibitem{alzubaidi2021review}
L.~Alzubaidi, J.~Zhang, A.~J. Humaidi, A.~Al-Dujaili, Y.~Duan, O.~Al-Shamma,
  J.~Santamar{\'\i}a, M.~A. Fadhel, M.~Al-Amidie, and L.~Farhan, ``Review of
  deep learning: Concepts, cnn architectures, challenges, applications, future
  directions,'' {\em Journal of big Data}, vol.~8, no.~1, pp.~1--74, 2021.

\bibitem{azulay2018deep}
A.~Azulay and Y.~Weiss, ``Why do deep convolutional networks generalize so
  poorly to small image transformations?,'' {\em arXiv preprint
  arXiv:1805.12177}, 2018.

\bibitem{zhang2019making}
R.~Zhang, ``Making convolutional networks shift-invariant again,'' in {\em
  International conference on machine learning}, pp.~7324--7334, PMLR, 2019.

\bibitem{sabri2014neural}
M.~Sabri, C.~Humphries, M.~Verber, E.~Liebenthal, J.~R. Binder, J.~Mangalathu,
  and A.~Desai, ``Neural effects of cognitive control load on auditory
  selective attention,'' {\em Neuropsychologia}, vol.~61, pp.~269--279, 2014.

\bibitem{wang2021taking}
J.~Wang, T.~Grant, S.~Velipasalar, B.~Geng, and L.~Hirshfield, ``Taking a
  deeper look at the brain: Predicting visual perceptual and working memory
  load from high-density fnirs data,'' {\em IEEE Journal of Biomedical and
  Health Informatics}, vol.~26, no.~5, pp.~2308--2319, 2021.

\bibitem{huang2021tufts}
Z.~Huang, L.~Wang, G.~Blaney, C.~Slaughter, D.~McKeon, Z.~Zhou, R.~Jacob, and
  M.~C. Hughes, ``The tufts fnirs mental workload dataset \& benchmark for
  brain-computer interfaces that generalize,'' 2021.

\bibitem{tolstikhin2021mlp}
I.~O. Tolstikhin, N.~Houlsby, A.~Kolesnikov, L.~Beyer, X.~Zhai, T.~Unterthiner,
  J.~Yung, A.~Steiner, D.~Keysers, J.~Uszkoreit, {\em et~al.}, ``Mlp-mixer: An
  all-mlp architecture for vision,'' {\em Advances in Neural Information
  Processing Systems}, vol.~34, pp.~24261--24272, 2021.

\bibitem{yucel2021best}
M.~A. Y{\"u}cel, A.~v. L{\"u}hmann, F.~Scholkmann, J.~Gervain, I.~Dan, H.~Ayaz,
  D.~Boas, R.~J. Cooper, J.~Culver, C.~E. Elwell, {\em et~al.}, ``Best
  practices for fnirs publications,'' {\em Neurophotonics}, vol.~8, no.~1,
  p.~012101, 2021.

\bibitem{lyu2021domain}
B.~Lyu, T.~Pham, G.~Blaney, Z.~Haga, A.~Sassaroli, S.~Fantini, and S.~Aeron,
  ``Domain adaptation for robust workload level alignment between sessions and
  subjects using fnirs,'' {\em Journal of Biomedical Optics}, vol.~26, no.~2,
  p.~022908, 2021.

\bibitem{zhong2020eeg}
P.~Zhong, D.~Wang, and C.~Miao, ``Eeg-based emotion recognition using
  regularized graph neural networks,'' {\em IEEE Transactions on Affective
  Computing}, vol.~13, no.~3, pp.~1290--1301, 2020.

\bibitem{schirrmeister2017deep}
R.~T. Schirrmeister, J.~T. Springenberg, L.~D.~J. Fiederer, M.~Glasstetter,
  K.~Eggensperger, M.~Tangermann, F.~Hutter, W.~Burgard, and T.~Ball, ``Deep
  learning with convolutional neural networks for eeg decoding and
  visualization,'' {\em Human brain mapping}, vol.~38, no.~11, pp.~5391--5420,
  2017.

\bibitem{yerkes1908relation}
R.~M. Yerkes, J.~D. Dodson, {\em et~al.}, ``The relation of strength of
  stimulus to rapidity of habit-formation,'' 1908.

\bibitem{lawhern2018eegnet}
V.~J. Lawhern, A.~J. Solon, N.~R. Waytowich, S.~M. Gordon, C.~P. Hung, and
  B.~J. Lance, ``Eegnet: a compact convolutional neural network for eeg-based
  brain--computer interfaces,'' {\em Journal of neural engineering}, vol.~15,
  no.~5, p.~056013, 2018.

\bibitem{boas2014twenty}
D.~A. Boas, C.~E. Elwell, M.~Ferrari, and G.~Taga, ``Twenty years of functional
  near-infrared spectroscopy: introduction for the special issue,'' 2014.

\bibitem{parasuraman2008neuroergonomics}
R.~Parasuraman and M.~Rizzo, {\em Neuroergonomics: The brain at work}.
\newblock Oxford University Press, 2008.

\bibitem{blaney2020design}
G.~Blaney, A.~Sassaroli, and S.~Fantini, ``Design of a source--detector array
  for dual-slope diffuse optical imaging,'' {\em Review of Scientific
  Instruments}, vol.~91, no.~9, p.~093702, 2020.

\bibitem{blaney2020phase}
G.~Blaney, A.~Sassaroli, T.~Pham, C.~Fernandez, and S.~Fantini, ``Phase
  dual-slopes in frequency-domain near-infrared spectroscopy for enhanced
  sensitivity to brain tissue: First applications to human subjects,'' {\em
  Journal of Biophotonics}, vol.~13, no.~1, p.~e201960018, 2020.

\bibitem{aghajani2017measuring}
H.~Aghajani, M.~Garbey, and A.~Omurtag, ``Measuring mental workload with eeg+
  fnirs,'' {\em Frontiers in human neuroscience}, vol.~11, p.~359, 2017.

\bibitem{hong2015classification}
K.-S. Hong, N.~Naseer, and Y.-H. Kim, ``Classification of prefrontal and motor
  cortex signals for three-class fnirs--bci,'' {\em Neuroscience letters},
  vol.~587, pp.~87--92, 2015.

\bibitem{lee2014hybrid}
M.-H. Lee, S.~Fazli, J.~Mehnert, and S.-W. Lee, ``Hybrid brain-computer
  interface based on eeg and nirs modalities,'' in {\em 2014 international
  winter workshop on brain-computer interface (BCI)}, pp.~1--2, IEEE, 2014.

\bibitem{benerradi2019exploring}
J.~Benerradi, H.~A.~Maior, A.~Marinescu, J.~Clos, and M.~L.~Wilson, ``Exploring
  machine learning approaches for classifying mental workload using fnirs data
  from hci tasks,'' in {\em Proceedings of the Halfway to the Future Symposium
  2019}, pp.~1--11, 2019.

\bibitem{garcia2003support}
G.~N. Garcia, T.~Ebrahimi, and J.-M. Vesin, ``Support vector eeg classification
  in the fourier and time-frequency correlation domains,'' in {\em First
  International IEEE EMBS Conference on Neural Engineering, 2003. Conference
  Proceedings.}, pp.~591--594, IEEE, 2003.

\bibitem{chollet2017xception}
F.~Chollet, ``Xception: Deep learning with depthwise separable convolutions,''
  in {\em Proceedings of the IEEE conference on computer vision and pattern
  recognition}, pp.~1251--1258, 2017.

\bibitem{yu2019review}
Y.~Yu, X.~Si, C.~Hu, and J.~Zhang, ``A review of recurrent neural networks:
  Lstm cells and network architectures,'' {\em Neural computation}, vol.~31,
  no.~7, pp.~1235--1270, 2019.

\bibitem{hochreiter1997long}
S.~Hochreiter and J.~Schmidhuber, ``Long short-term memory,'' {\em Neural
  computation}, vol.~9, no.~8, pp.~1735--1780, 1997.

\bibitem{graves2014neural}
A.~Graves, G.~Wayne, and I.~Danihelka, ``Neural turing machines,'' {\em arXiv
  preprint arXiv:1410.5401}, 2014.

\bibitem{sommer2021classification}
N.~M. Sommer, B.~Kakillioglu, T.~Grant, S.~Velipasalar, and L.~Hirshfield,
  ``Classification of fnirs finger tapping data with multi-labeling and deep
  learning,'' {\em IEEE Sensors Journal}, vol.~21, no.~21, pp.~24558--24569,
  2021.

\bibitem{chung2014empirical}
J.~Chung, C.~Gulcehre, K.~Cho, and Y.~Bengio, ``Empirical evaluation of gated
  recurrent neural networks on sequence modeling,'' {\em arXiv preprint
  arXiv:1412.3555}, 2014.

\bibitem{lee2020self}
H.~Lee, S.~J. Hwang, and J.~Shin, ``Self-supervised label augmentation via
  input transformations,'' in {\em International Conference on Machine
  Learning}, pp.~5714--5724, PMLR, 2020.

\bibitem{arjovsky2017wasserstein}
M.~Arjovsky, S.~Chintala, and L.~Bottou, ``Wasserstein generative adversarial
  networks,'' in {\em International conference on machine learning},
  pp.~214--223, PMLR, 2017.

\bibitem{sejdinovic2013equivalence}
D.~Sejdinovic, B.~Sriperumbudur, A.~Gretton, and K.~Fukumizu, ``Equivalence of
  distance-based and rkhs-based statistics in hypothesis testing,'' {\em The
  annals of statistics}, pp.~2263--2291, 2013.

\bibitem{kang2019contrastive}
G.~Kang, L.~Jiang, Y.~Yang, and A.~G. Hauptmann, ``Contrastive adaptation
  network for unsupervised domain adaptation,'' in {\em Proceedings of the
  IEEE/CVF conference on computer vision and pattern recognition},
  pp.~4893--4902, 2019.

\bibitem{ajakan2014domain}
H.~Ajakan, P.~Germain, H.~Larochelle, F.~Laviolette, and M.~Marchand,
  ``Domain-adversarial neural networks,'' {\em arXiv preprint arXiv:1412.4446},
  2014.

\bibitem{hendrycks2016gaussian}
D.~Hendrycks and K.~Gimpel, ``Gaussian error linear units (gelus),'' {\em arXiv
  preprint arXiv:1606.08415}, 2016.

\bibitem{he2016deep}
K.~He, X.~Zhang, S.~Ren, and J.~Sun, ``Deep residual learning for image
  recognition,'' in {\em Proceedings of the IEEE conference on computer vision
  and pattern recognition}, pp.~770--778, 2016.

\bibitem{bak2019open}
S.~Bak, J.~Park, J.~Shin, and J.~Jeong, ``Open-access fnirs dataset for
  classification of unilateral finger-and foot-tapping,'' {\em Electronics},
  vol.~8, no.~12, p.~1486, 2019.

\bibitem{kocsis2006modified}
L.~Kocsis, P.~Herman, and A.~Eke, ``The modified beer--lambert law revisited,''
  {\em Physics in Medicine \& Biology}, vol.~51, no.~5, p.~N91, 2006.

\bibitem{kingma2014adam}
D.~P. Kingma and J.~Ba, ``Adam: A method for stochastic optimization,'' {\em
  arXiv preprint arXiv:1412.6980}, 2014.

\bibitem{oostenveld2001five}
R.~Oostenveld and P.~Praamstra, ``The five percent electrode system for
  high-resolution eeg and erp measurements,'' {\em Clinical neurophysiology},
  vol.~112, no.~4, pp.~713--719, 2001.

\end{thebibliography}

\end{document}